\pdfoutput=1
\documentclass[runningheads]{llncs}
\usepackage{graphicx}
\usepackage{subcaption}
\usepackage{multirow}
\usepackage{hyperref}
\hypersetup{colorlinks,allcolors=black}
\usepackage{comment}
\usepackage{amsmath,amsbsy,amsfonts,amssymb,units,bm}
\usepackage[mathscr]{eucal}
\newcommand{\partitle}[1]{\smallskip \noindent \textbf{#1.}}

\usepackage[linesnumbered,ruled,vlined]{algorithm2e}

\SetKwInput{KwInput}{Input}
\SetKwInput{KwOutput}{Output}

\begin{document}
%
\title{RobustFed: A Truth Inference Approach for Robust Federated Learning}
\titlerunning{RobustFed: A truth inference approach for robust federated learning}
%
\author{Farnaz Tahmasebian \and Jian Lou \and
 Li Xiong 
\institute{Emory University, Computer Science \\
 \email{\{ftahmas,jian.lou,lxiong\}@emory.edu}\\
 }
 }


%

%
\maketitle 
\begin{abstract}
Federated learning is a prominent framework that enables
clients (e.g., mobile devices or organizations) to train a collaboratively global model under a central server's orchestration while keeping local training datasets' privacy. However, the aggregation step in federated learning is vulnerable to adversarial attacks as the central server cannot manage clients' behavior. Therefore, the global model's performance and convergence of the training process will be affected under such attacks. To mitigate this vulnerability issue, we propose a novel robust aggregation algorithm inspired by the truth inference methods in crowdsourcing via incorporating the worker's reliability into aggregation. We evaluate our solution on three real-world datasets with a variety of machine learning models. Experimental results show that our solution ensures robust federated learning and is resilient to various types of attacks, including noisy data attacks, Byzantine attacks, and label flipping attacks.

\end{abstract}
\section{Introduction}
Federated learning (FL) has emerged as a promising new collaborative learning framework to build a shared model across multiple clients (e.g., devices or organizations) while keeping the clients’ data private~\cite{mcmahan2017communication,lim2020federated,abadi2016deep}. The latter is also known as cross-silo FL, which we focus on in this paper. Such a framework is practical and flexible and can be applied in various domains, such as conversational AI and healthcare~\cite{mcmahan2017communication,mothukuri2020survey,lim2020federated}. Training a generalizable model for these domains requires a diverse dataset. Accessing and obtaining data from multiple organizations and centralizing them in a third-party service provider can be impractical considering data privacy concerns or regulations. Yet, we still wish to use data across various organizations because a model trained on data from one organization may be subject to bias and poor generalization performance. FL makes it possible to harness the data for joint model training with better generalization performance without the requirement to share raw private local datasets~\cite{abadi2016deep}.

\begin{figure}[]
\centering
\includegraphics[width=0.9\textwidth]{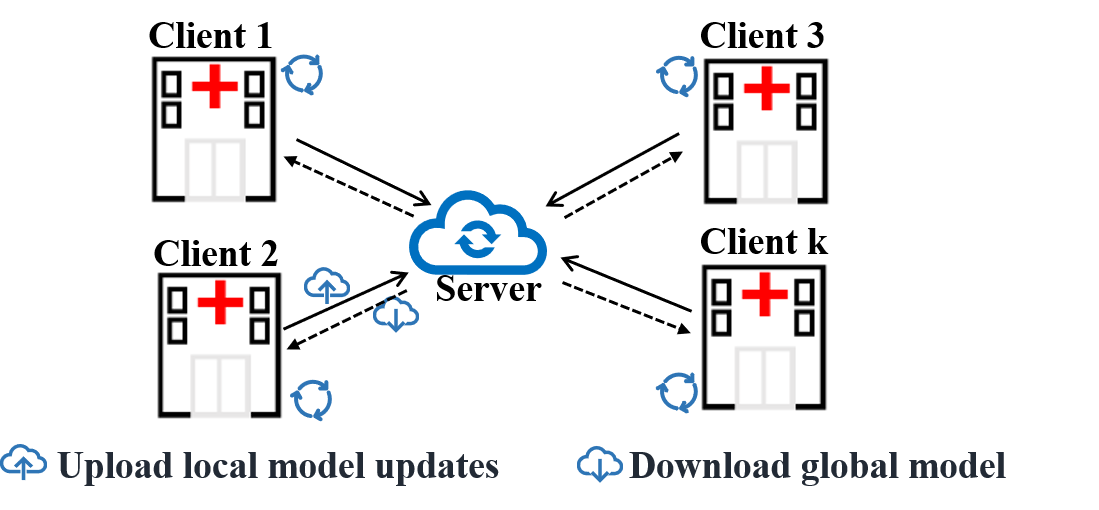}
\vspace{-1em}
\caption{Overview of Cross-silo Federated Learning (FL) Framework}
\label{fig:FL_Overview}
\end{figure}

In a cross-silo FL framework (as shown in Figure~\ref{fig:FL_Overview}), there is a semi-honest global coordinating server and several participating clients. The global server controls the learning process and aggregates the model parameters submitted by clients during multiple communication rounds. The clients train the same model locally using their local datasets. Then, they share their updated local model parameters, not their raw data, with the server, which aggregates all their contributions and broadcasts back the updated global model parameters. 

The most commonly used aggregation algorithm is called Federated Averaging (FedAvg)~\cite{mcmahan2017communication} that takes a weighted average of the local model parameters. This aggregation method is vulnerable to adversarial attacks or unintentional errors in a system. Due to strategic adversarial behavior (e.g., label-flipping and Gaussian noise attacks~\cite{kairouz2019advances,damaskinos2019aggregathor,gu2019badnets,blanchard2017machine}) or infrastructure failures (e.g., Byzantine faults~\cite{lamport2019byzantine} where client nodes act arbitrarily), the clients can send malicious (manipulated) or arbitrary values to the server. Thus, the global model can be affected severely. Therefore, robust FL against such potential behaviors or failures is essential. 

Recently, several methods have been proposed to mitigate attacks in FL or distributed learning~\cite{fung2020limitations,chen2017distributed,yin2018byzantine,fung2018mitigating,blanchard2017machine}. The statistical methods such as median or trimmed mean based aggregation (instead of weighted averaging) ~\cite{yin2018byzantine} perform well under Byzantine attack. However, they fail under other types of attacks such as label-flipping and Gaussian noise attacks. 

This paper proposes using a truth inference approach for robust aggregation against such attacks in FL. Truth inference is a key component of crowdsourcing that aggregates the answers of the crowd (i.e., workers) to infer the true label of tasks (e.g., traffic incidents, image annotation)~\cite{raykar2010learning,karger2011iterative}. We make this connection for the first time that the model parameter aggregation can be formulated as a truth inference problem, i.e., each client is a worker, the local parameters (answers) by the workers need to be aggregated to estimate the global parameter (label). The key idea is to explicitly model the reliability of clients and take them into consideration during aggregation. Such an approach has shown promising results in crowdsourcing compared to simple aggregation approaches such as majority voting (or averaging). However, there are several challenges and opportunities in applying the truth inference approach for robust FL (compared to crowdsourcing). First, an attacker can manipulate the local training data (e.g., adding noise or flipping the labels) to affect the model parameters (versus directly changing the model parameters). The server only observes the model parameters without access to the data. Hence, a direct application of the truth inference approach on the model parameters cannot detect the malicious clients reliably. Second, FL requires multi-round communication of the local model parameters to the server. This dynamic information creates both challenges and opportunities in detecting unreliable clients. Finally, as in many practical settings, the server does not have access to any golden validation set for validating the local parameter models in order to detect unreliable clients.



To address these challenges, we derive the clients' reliability score by solving an optimization problem over multiple iterations of FL. We then incorporate the reliability of each client in the aggregation. Our approach is based on two main insights. First, the existing truth inference approaches rely entirely on the derived reliability of the workers for aggregation. In our case, since the model parameters may not accurately reflect the reliability of the workers due to the different kinds of attacks (e.g., label-flipping), we use a pruning algorithm that removes clients with outlier reliability, which mitigates the impact of the malicious clients during aggregation. 
Second, we exploit the multi-round model parameters submitted by the clients for evaluating the client's reliability in a more robust way. We briefly summarize our contributions as follows.

\begin{itemize}
 \item We develop a novel robust aggregation method for FL against potential adversarial attacks and Byzantine failures of clients. The method explicitly models the clients' reliability based on their submitted local model parameters and incorporates them into aggregation, hence providing a robust estimate of the global model parameters. 
 \item We further enhance the aggregation method by exploiting the multi-round communication of FL and considering the model parameters submitted by the clients both in the previous rounds and the current round for evaluating the client's reliability. 
 \item We compare our proposed method to several baselines on three image datasets. The results show that our proposed aggregation methods mitigate the impact of attacks and outperform other baselines.
\end{itemize}

\section{Related Works}
In this section, we provide a brief review of adversarial attacks on federated learning (FL) along with the existing defense and robustness methods in FL. Subsequently, we briefly review truth inference methods in crowdsourcing. 

\subsection{Adversarial Attacks on Federated Learning}
In federated learning (FL), all the participants agree on a common learning objective and model structure. The attacker aims to compromise the global model by uploading the malicious data to the global server~\cite{mcmahan2017communication}. The adversary can control the whole local training dataset, local hyper-parameter of a model, and local model parameters in this system.  

This paper mainly considers the data poisoning attack scenario, in which malicious clients create poisoned training samples and inject them into their local training dataset~\cite{fung2018mitigating}. Then, the local model is trained on the dataset contaminated with such poisoned samples. The purpose of this attack is to manipulate the global model to misclassify on test datasets. These attacks can be further divided into two categories: 1) label-flipping attacks~\cite{fung2018mitigating} and 2) noisy features attack~\cite{fung2018mitigating}. The label-flipping attack occurs where the labels of training examples of one class are flipped to another class while the data features remain unchanged. For example, an attacker can train a local model with cat images misclassified as a dog and then share the poisoned local model for aggregation. A successful attack forces a model to incorrectly predicts cats to be dogs. In the noisy features attacks, the adversary adds noise to the features while keeping the class label of each data point intact~\cite{fung2018mitigating}. Noisy data and the backdoor attacks fall in this type of attack~\cite{xie2019dba,wang2019neural}. 

FL is vulnerable to poisoning attacks. Studies~\cite{fung2018mitigating,bhagoji2019analyzing} show that just one or two adversarial clients are enough to compromise the performance of the global model. Thus, developing a robust method against these attacks is essential. Fung et al.~\cite{fung2018mitigating} proposed a defense method, called FoolsGold, against data poisoning attack in FL in a non-IID setting. Their solution differentiates the benign clients from the adversary ones by calculating the similarity of their submitted gradients. Other techniques use the recursive Bayes filtering method~\cite{munoz2019byzantine} to mitigate the data poisoning attack. 
In some studies~\cite{bhagoji2019analyzing,sattler2019robust}, researchers assume that the global server has access to a golden validation dataset that represents data distribution from clients. The server can detect adversaries by assessing the effectiveness of provided updates on the global model's performance. If the updates do not improve the global model's performance, the client is flagged as a potential adversary~\cite{bhagoji2019analyzing}. However, this method requires the validation dataset which is difficult to achieve in practice.

\subsection{Byzantine-Robust Federated Learning}
Byzantine clients aim to prevent the global model's convergence or lead the global model to converge to a poor solution. In some scenarios, the Byzantine clients choose to add Gaussian noise to the gradient estimators, then send these perturbed values to the server. The Byzantine gradients can be hard to distinguish from the benign clients since their variance and magnitude are similar to the benign gradient submissions. Byzantine-Robust methods have been studied in recent years~\cite{alistarh2018byzantine,yin2018byzantine,munoz2019byzantine,guerraoui2018hidden,blanchard2017machine,li2019rsa,chen2017distributed}. Most existing methods assume that data is distributed IID among clients and are based on robust statistical aggregation.

A common aggregation method against the Byzantine attack is based on the median of the updates~\cite{chen2017distributed}. This method aggregates each model parameter independently. It sorts the local models' jth parameters and takes the median as the jth parameter for the global model. Trimmed mean~\cite{yin2018byzantine} is another method that sorts jth parameters of all local models, then removes the largest and smallest of them, and computes the mean of the remaining parameters as the jth parameter of the global model. 
Krum~\cite{blanchard2017machine} selects one of the local models that are similar to other models as the global model. Krum first computes the nearest neighbors to each local model. Then, it calculates the sum of the distance between each client and their closest local models. Finally, select the local model with the smallest sum of distance as the global model. Aggregation methods such as Krum and trimmed mean need to know the upper bound of the number of compromised workers. Other methods extend Krum, such as Multi-Krum~\cite{blanchard2017machine} and Bulyan~\cite{guerraoui2018hidden}. Multi-Krum combines Krum and averaging. Bulyan combines Krum and trimmed mean. It iteratively applies Krum to local models then applies trimmed mean to aggregate the local models.

\subsection{Truth Inference Methods}
Crowdsourcing aggregates the crowd's wisdom (i.e., workers) to infer the truth label of tasks in the system, which is called truth inference. Effective truth inference, especially given sparse data, requires assessment of workers' reliability. There exist various approaches to infer the truth of tasks~\cite{jagabathula2014reputation,li2014resolving,dawid1979maximum,venanzi2014community,kim2012bayesian,gaunt2016training,zheng2017truth}, including direct computing~\cite{jagabathula2014reputation}, optimization~\cite{jagabathula2014reputation,li2014resolving}, probabilistic graphical model (PGM)~\cite{dawid1979maximum,venanzi2014community,kim2012bayesian}, and neural network based~\cite{liaggregating}. The simplest method is majority voting, which works well if all workers provide answers to all of the tasks. However, it fails when data is sparse and workers may be unreliable, as in many practical settings. 

Recently, two experimental studies compared state-of-the-art truth inference methods in a "normal" setting and "adversarial" setting~\cite{zheng2017truth,tahmasebian2020crowdsourcing}. The "adversarial" environment is where workers intentionally or strategically manipulate the answers. In the "normal" setting, the study~\cite{zheng2017truth} concluded that truth inference methods that utilize a PGM have the best performances in most settings where the type of tasks are binary and single label. The study in the "adversarial" settings~\cite{tahmasebian2020crowdsourcing} focusing on binary tasks showed that neural networks and PGM based methods are generally more robust than other methods for the binary type of tasks. 
In our FL setting, since we are dealing with model parameters that are numeric and updates that are dense (i.e. a subset of participants submit their model parameters in each round), we use an optimization based truth inference method PM as a baseline method. 

\section{Preliminaries}
\subsection{Federated Learning (FL)}
The FL framework is important when the participating organizations desire to keep their data private. Instead of sharing data, they share the model parameters to take advantage of a high volume of data with different distributions and improve the model's generalization. FL consists of $K$ clients and a global server $G$. 
Each client $c_i$ has their own local dataset \textbf{D}$_{i}$ = $\{x_1^i,....x_{l_i} ^ i\}$, where $|D_i|=l_i$. The total number of samples across all the clients is $\sum_{i=1}^{K} l_i = l$. The goal of FL is to keep the data local and learn a global model with $n$ parameters $w_G \in \mathbb{R}^n$ which minimizes the loss among all samples $D= \bigcup_{i=1}^{K} D_i$ in the aim that the model generalizes well over the test data \textbf{D}$_{test}$.

At each time step $t$, a random subset from the clients is chosen for synchronous aggregation, i.e. the global server computes the aggregated model, then sends the latest update of the model to all selected clients. Each client $c_i\in {K}$ uses their local data \textbf{D}$_{i}$ to train the model locally and minimize the loss over its own local data. After receiving the latest global model, the clients starts the new round from the global weight vector $w_G ^t$ and run model for E epochs with a mini-batch size B. At the end of each round, each client obtains a local weight vector $w_{c_i}^{t+1}$ and computes its local update $\delta_{c_i} ^ {t+1}$= $w_{c_i}^{t+1} - w_{G}^{t}$, then sends the corresponding local updates to the global server, which updates the model according to a defined aggregation rule. The simplest aggregation rule is a weighted average, i.e., Federated Averaging (FedAvg), and formulated as follow, where $\alpha_i =\frac{l_i}{l}$ and $\sum_{i=1}^{K} \alpha_i= 1$. 
\begin{equation}
w_{G} ^{t+1} = w_{G}^{t} + \sum_{i=1}^{K} \alpha_i \cdot \delta_i ^{t+1}
\end{equation}

\subsection{Adversarial Model}
We assume any of the clients can be attackers who have full access to the local training data, model structure, learning algorithms, hyperparameters, and model parameters. The adversary's goal is to ensure the system's performance degrades or causes the global model to converge to a bad minimum. 

In this paper, we mainly consider the data poisoning attack and Byzantine attack. The data poisoning attack is applied in the local training phase and divided into label-flipping and noisy data attacks. In each round, the attacker trains a new local model (based on the global model from the previous round) on the poisoned training data and uploads the new model parameters to the server. Byzantine attack directly changes the model parameters to be uploaded to the server. For the adversarial model, we follow two assumptions: (1) The number of adversaries is less than 50\% of whole clients; (2) the data is distributed among the clients in an independent and identically (IID) fashion. 

\section{Proposed Robust Model Aggregation}
\begin{figure}[]
\centering
\includegraphics[width=\textwidth]{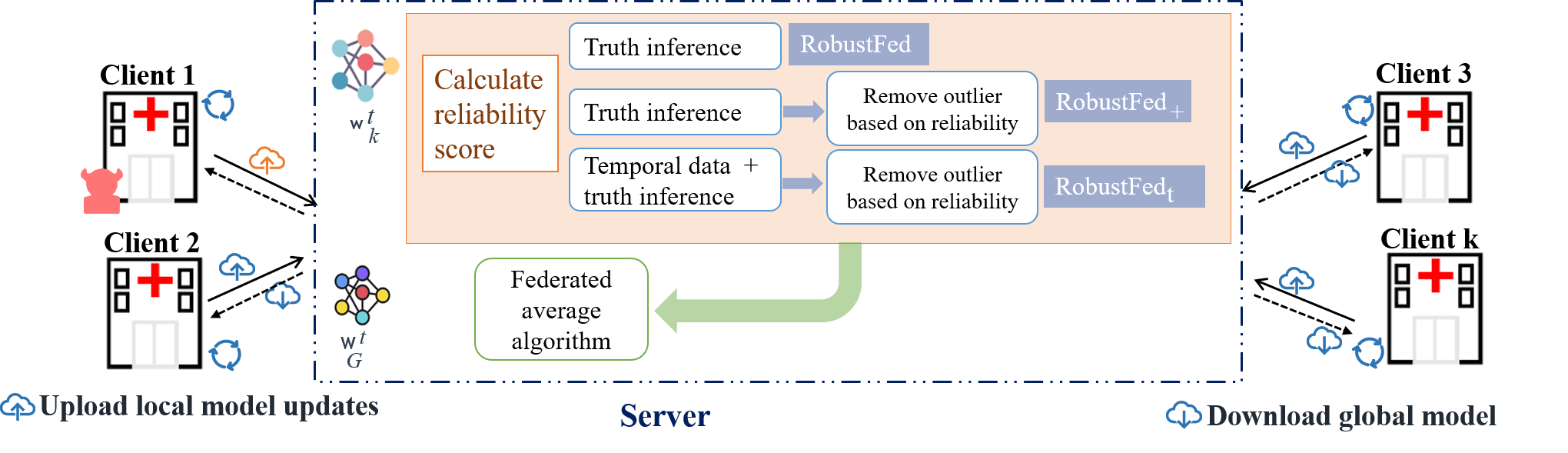}
\vspace{-1em}
\caption{Overview of Proposed Methods}
\label{fig:system_model}
\end{figure}

We present our proposed robust aggregation method in this section. The key idea is to explicitly model the reliability of clients inspired by truth inference algorithms and take them into consideration during aggregation. We first introduce the truth inference framework and utilize it in FL to estimate the reliability of provided updates by clients in each round. We further improve it by removing the outlier clients before aggregation to address its limitations of correctly detecting malicious clients in data poisoning attacks. Finally, we incorporate the multi-round historical model parameters submitted by the clients for more robust aggregation. 
The high-level system model is illustrated in Figure~\ref{fig:system_model}. The server comprises two modules: (1) the reliability score calculator; and (2) the aggregator. The server calculates each client's reliability based on three proposed methods that is improved upon each other. 

\subsection{Truth Inference Method}
\label{sec:truth}
Due to the openness of crowdsourcing, the crowd may provide low-quality or even noisy answers. Thus, it is crucial to control crowdsourcing's quality by assigning each task to multiple workers and aggregating the answers given by different workers to infer each task's correct response. 
The goal of truth inference is to determine the true answer based on all the workers' answers for each task. 

\begin{figure}[]
\centering
\includegraphics[width=0.85\textwidth]{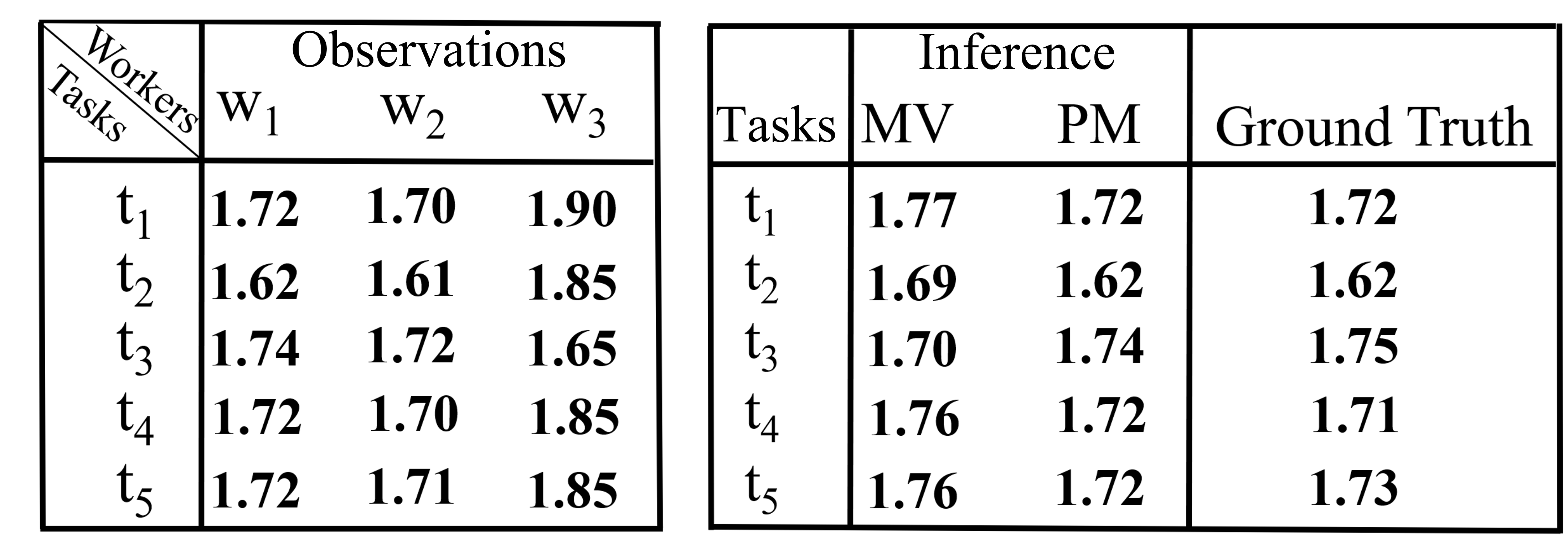}
\caption{Example of Crowdsourcing System}
\label{fig:ex_crowd_pm}
\end{figure}%

Figure~\ref{fig:ex_crowd_pm} shows an example given three workers \textbf{W}=\{$w_1,w_2,w_3$\} and five tasks \textbf{T}=\{$t_1,t_2,..,t_5$\}, the goal is to infer the true answer for each tasks. For example, worker $w_1$ provides 1.72 as an answer to task $t_4$. A naive solution to infer the true answer per task is Majority Voting (MV) or averaging. Based on Figure~\ref{fig:ex_crowd_pm}, the truth derived by MV for task $t_1$ is 1.77, which is inferred incorrectly. A more advanced method such as PM~\cite{li2014resolving} models the reliability of each worker explicitly and resolves conflicts from different sources for each entry. Compared with the ground truth answers, it is clear that worker $w_1$ and $w_2$ provide more accurate information (more reliable) while $w_3$ is not very reliable. By modeling and learning the reliability of workers, PM provides more accurate results compared with averaging.

We can map the model aggregation at the server in FL into the truth inference problem by considering the model's weight parameters as tasks. 
In both crowdsourcing and FL, we deal with unlabeled data. In crowdsourcing, the true label of tasks are not available; in FL, the true parameters of the model are unknown (the server does not have access to any validation dataset). The parameter aggregation can be considered as a numeric task (as versus binary task).
Algorithm~\ref{alg:get_reliability} shows the truth inference framework for numeric tasks. The reliability of each worker $i \in [k]$ is denoted as $r_{c_i}$. It initializes clients’ reliability with the same reliability as $r_{c_i} =1$. Also, it initializes the estimated truth for each weight parameter as the median of all values provided by the clients. Then it adopts an iterative approach with two steps, 1) inferring the truth and 2) estimating client reliability. 

\begin{algorithm}[]
\label{alg:get_reliability}
\caption{Obtain Clients Reliability}
\footnotesize
\SetAlgoLined
\DontPrintSemicolon
 \KwInput{Provided parameters by local clients $\bm{\delta}_{k} = \bigcup_{i=1}^{K} \delta_{c_i}, w_G^t$ }
 
 \KwOutput{R=$\bigcup_{i=1}^{K} r_{c_i}$}
 
 Initialize clients’ reliability ($r_{c_i}=1~  for ~ i \in K$)
 
 Initialize inferred truth of each update parameter ($\hat{\Delta}_G$) as the median of local updates of $\bm{\delta}_{k}$

 \While{True}{
 // Step 1: Inferring the Truth\;
 \For{each weight parameter $j \in N$}{
 Inferring the $\hat{\Delta}_G$ based on $\bm{\delta}_{k}$ and $\textbf{R}$
 }
 // Step 2: Estimating client reliability

 \For{each client}
 {estimate $\textbf{R}$ based on $\bm{\delta}_{k}$ and $\hat{\Delta}_G$}
 \If{converge}
 {break\;}
 
 }
\end{algorithm}

\subsection{Robust Aggregation Method: RobustFed}
In this section, details of our proposed aggregation method are provided. To begin each round, we compute the reliability level of each client by applying the truth inference method. 

Let $\bm{\delta}_{c_i}^t=\{\bm{\delta}_{c_i}^t[1],\bm{\delta}_{c_i}^t[2],...,\bm{\delta}_{c_i}^t[n]\}$ be the local updates that is shared by client $c_i$ at round $t$. Let $\mathcal{K}= \{c_1,c_2,...c_k\}$ be the set of clients. Hence, at round $t$, the updated parameters $\bm{\delta}_{k}^t$ are collected from $K$ clients. Given the updated parameters $\bm{\delta}_k^t$ provided by $K$ clients, the goal of utilizing the truth inference is to infer the reliability of each clients $\bm{R} = \{r_{c_1}, ...r_{c_k}\}$ and incorporate this reliability score into the aggregation method.

The idea is that benign clients provide trustworthy local updates, so the aggregated updates should be close to benign clients' updates. Thus, we should minimize the weighted deviation from the true aggregated parameters where the weight reflects the reliability degree of clients. Based on this principle, we utilize the PM method, which is a truth inference method applicable in numerical tasks~\cite{li2014resolving}. First, by minimizing the objective function, the values for two sets of unknown variables $\Delta$ and $\bm{R}$, which correspond to the collection of truths and clients' reliabilities are calculated. The loss function measures the distance between the aggregated parameters (estimated truth) and the parameters provided by client (observation). When the observation deviates from the estimated truth, the loss function return a high value. To constrain the clients' reliabilities into a certain range, the regularization function is defined and it reflects the distributions of clients' reliabilities.

Intuitively, a reliable client is penalized more if their observation is quite different from the estimated truth. In contrast, the observation made by an unreliable client with low reliability is allowed to be further from the truth. To minimize the objective function, the estimated truth relies more on the clients with high reliability. The estimated truth and clients' reliabilities are learned together by optimizing the objective function through a joint procedure. We formulate this problem as an optimization problem as follows:

\begin{equation}
 \min_{\bm{R}, \hat{\bm{\Delta}}} \sum_{i=1}^{K} {r_{c_i}} \cdot~dist~(\hat{\Delta}_{G}, \bm{\delta}_{c_i}^t),
\label{eq:pm_method}
\end{equation}
 
where $r_{c_i}$, $\bm{\delta}_{c_i}^t$ and $\hat{\Delta}_{G}$ represent client $c_i$'s reliability, provided update by client $c_i$ at time $t$, and aggregated updates at time $t$ on the global server, respectively. Also $dist~(\hat{\Delta}_{G}, \bm{\delta}_{c_i}^t)$ is a distance function from the aggregated updates of all clients to the clients' provided update.
The goal is to minimize the overall weighted distance to the aggregation parameters in the global server in a way that reliable clients have higher weights (importance).
 In our problem, the type of parameters provided by clients are continuous, therefore Euclidean distance is used as a distance function, $\sqrt{\sum_{j=1}^{N}\left ( \bm{\hat{\Delta}_{G}^j} - \bm{\delta_{c_i}^j} \right )^2}$, where N is the number of local parameters and $\delta_{c_i}^j$ indicates the j-th local parameter shared by client $c_i$. The client $c_i$'s reliability is modeled using a single value $r_{c_i}$. Intuitively, workers with answers deviating from the inferred truth tend to be more malicious. The algorithm iteratively conducts the following two steps, 1) updating the client's reliability and 2) updating the estimated truth for parameters. 

To update the client's reliability, we fix the values for the truths and compute the clients' reliability that jointly minimizes the objective function subject to the regularization constraints. Initially, each client is assigned with the same reliability, $\forall_{i \in ~\mathcal{K}}$ $r_{c_i}$=1. The reliability score of each client after each iteration is updated as: 

\begin{equation}
\begin{aligned}
r_{c_i} = -\log \left ( \frac{ \sum_{j=1}^{N} dist({\hat{\Delta}_{G}}^j~,~\delta_{c_i}^j)} {\sum_{k'=c_{1}}^{c_{K}} \sum_{j=1}^{N} dist({\hat{\Delta}_{G}}^j~,~{\delta_{k^{'}}}^j)} \right )
\label{eq:pm_rel_client}
\end{aligned}
\end{equation}

Equation~\ref{eq:pm_rel_client} indicates that a clients reliability is inversely proportional to the difference between its observations and the truths at the log scale.

By fixing the reliability of clients, the truths of parameters are updated in a way that minimizes the difference between the truths and the client’s observations where clients are weighted by their reliabilities and calculated as: $\hat{\Delta}_{G} = \frac{\sum_{i=1}^{K} r_{c_i} \cdot \delta_{c_i}}  {\sum_{i=1}^{K} r_{c_i}}$

At the aggregation step, the global server incorporates the provided parameters of each clients based on their reliability. Hence, the global parameters are updated as follows: 

\begin{equation}
w_{G}^{t+1} = w_{G}^{t} + \sum_{i \in K} r_{c_i}^{t} \cdot \alpha_{i} \cdot \delta_{c_i}^{t+1}
\end{equation}

\subsection{Reduce Effect of Malicious Clients: RobustFed$_+$}

RobustFed incorporate the reliability of every client in the aggregation but does not include explicit mechanisms to detect and exclude malicious clients. To further reduce the effect of malicious clients, we further propose RobustFed$_+$ to detect non-reliable clients at each round and discard their participation during the aggregation phase. 

\begin{algorithm}[]
\DontPrintSemicolon
\footnotesize
 \KwInput{selected clients $K^t$,\, $\textbf{R}^{t}$ (reliability of all clients), $w_{G}^t$, }
 \KwOutput{$w_{G}^{t+1}$}
 
 $\textbf{Cand}$ (set of clients' candidate) initialized to  $\emptyset$
 
 $\textbf{R}^{t} \gets getClientsReliablity()$
 
 $\bar\mu, \sigma \gets median(\textbf{R}^{t}), std(\textbf{R}^{t})$
 
 \For {$i \in K$}
 {
 \If{ $\bar\mu -\sigma <= r_{c_i}^{t} <= \bar\mu+\sigma$}
 {
 Add $c_i$ ~to ~$\textbf{Cand}$ 
 }
 }
 
 $w_{G}^{t+1} \gets w_{G}^{t}~ + \sum_{i\in[\textbf{Cand}]} r_{c_i}^{t} \cdot \alpha_i \cdot \delta_{c_i}^{t+1}$
\caption{Robust Aggregation (RobustFed$_+$)}
\label{alg:robust_agg}
\end{algorithm}

Algorithm~\ref{alg:robust_agg} summarizes RobustFed$_+$ method. After obtaining the reliability of each clients, the median ($\bar\mu$) and standard deviation ($\sigma$) of the reliabilities are computed for all the clients participated in the round $t$. The clients whose reliability fit in the range of $[\bar\mu - \sigma, \bar\mu + \sigma]$ are selected as a candidate, and the global parameters are updated as follows: {$w_{G}^{t+1}$ = $w_{G}^{t}$~ + ~$\sum_{i\in [\bf{Cand}]} r_{c_i}^{t} \cdot \alpha_i \cdot \delta_{c_i}^{t+1}$}.

We note that a straightforward method is to remove the clients with lowest reliability scores. Intuitively, we expect the server to assign a higher reliability to honest clients and a lower score to the malicious ones. In our experimental studies, we indeed observe this when no attack happens or under specific types of attacks such as Byzantine or data noise attacks. However, under label-flipping attack, we observe that the RobustFed method assigns higher reliability to the malicious clients. This is because the gradients of the malicious clients can be outliers under such attacks and significantly dominates (biases) the aggregated model parameters, and hence has a high reliability because of its similarity to the aggregated values. Therefore, in our approach, we disregard the clients with reliability deviating significantly from the others. 

\subsection{Incorporate the Temporal Data to Improve the Defense Capability: RobustFed$_t$}
Given the multi-round communication between the clients and the server in FL, RobustFed and RobustFed$_+$ only consider one round and ignore the temporal relationship among weight parameters in multiple rounds. Ignoring this temporal relationship might miss important insights of the parameters shared by clients at each rounds. Intuitively, under data poisoning or label flipping attacks, considering the parameters over multiple rounds will more effectively reveal the malicious clients. To take advantage of temporal information, we propose RobustFed$_t$ to incorporate the statistical information of the previous rounds during the reliability estimation. Incorporating the statistical information is dependent on the way the clients are selected in each round:\\
\textbf{Static Setting:} The server selects the same set of clients at each round to participate in training global model. Therefore, we add the statistics of the model parameters from previous rounds as new tasks in addition to the vector of weights. These statistics are the number of large weights, number of small weights, median of weights and average of weights. The reliability is then evaluated based on all statistics and the parameters submitted in current rounds.\\
\textbf{Dynamic Setting}: The server dynamically selects a set of clients to join FL and participate in training global model. Since each client may participate with different frequency, we only add median and average of weights from previous round as the weights provided by the new clients.

\section{Evaluation}
\vspace{-0.5 em}

\subsection{Experiment Settings}

\subsubsection{Dataset.} We consider the following three public datasets.
\vspace{-0.5em}
\begin{itemize}
 \item MNIST dataset: This dataset contains 70,000 real-world hand written images with digits from 0 to 9 with 784 features. We split this dataset into a training set and test set with 60,000 and 10,000 samples respectively. 
 \item Fashion-MNIST (fMNIST) dataset: This dataset consists of 28×28 gray scale images of clothing and footwear items with 10 type of classes. The number of features for this dataset is 784. We split this dataset in which training has 60,000 and test data has 10,000 samples. 
 \item CIFAR-10 dataset: This dataset contains 60,000 natural color image of 32x32 pixels in ten object classes with 3,072 features. We split this dataset in which training has 50,000 and test data has 10,000 samples. 
\end{itemize}

For MNIST and fMNIST datasets, we use a 3-layer convolutional neural network with dropout (0.5) as the model architecture. The learning rate and momentum are set as 0.1 and 0.9, respectively. 
For CIFAR-10, we use VGG-11 as our model. The droput, learning rate and momentum are set as 0.5, 0.001, 0.9, respectively. 

\vspace{-2.em}
\subsubsection{Experiment Setup and Adversarial Attacks.} 
We consider the training data split equally across all clients. For selecting clients to participate in each round, two selection methods are considered, 1) static mode and 2) dynamic mode. In the static mode, the number of clients are set to be 10 and at each iteration, the same set of clients are chosen. In the dynamic mode, the server randomly selects 10 clients from the pool of 100 clients in each round. 

We assume that 30\% of the clients are adversary. We consider three attack scenarios. 
\vspace{-0.5em}
\begin{itemize}
 \item {Label-Flipping Attacks:} Adversaries flip the labels of all local training data on one specific class (e.g., class \#1) and train their models accordingly. 

 \item{Noisy Data:} In MNIST and FMNIST, the inputs are normalized to the interval [0,1]. In this scenario, for the selected malicious clients, we added uniform noise to all the pixels, so that $x \gets x+ U$(-1.4,1.4). Then we cropped the resulting values back to the interval [0,1]. 
 
 \item{Byzantine Attack:} Adversary perturb the model updates and send the noisy parameters to the global server. $\delta_{i}^{t} \gets \delta_{i}^{t} + \epsilon$, where $\epsilon$ is a random perturbation drawn from a Gaussian distribution with $\mu = 0$ and $\sigma = 20$.

\end{itemize}


\begin{figure}[]
\begin{subfigure}{0.28\textwidth}
\includegraphics[width=\textwidth]{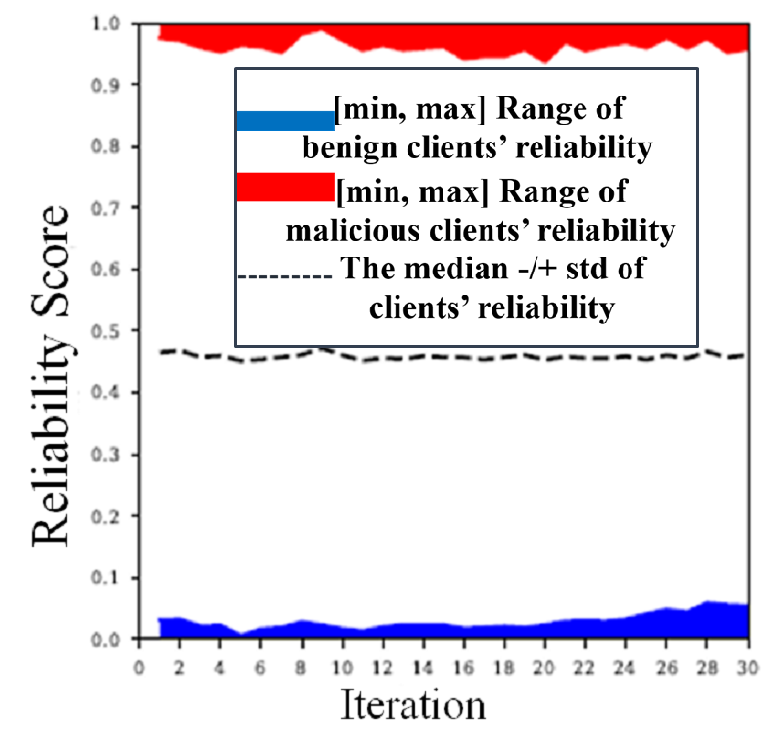}
\caption{Flipping($RobustFed$)}
\end{subfigure}%
\begin{subfigure}{0.28\textwidth}
\includegraphics[width=\textwidth]{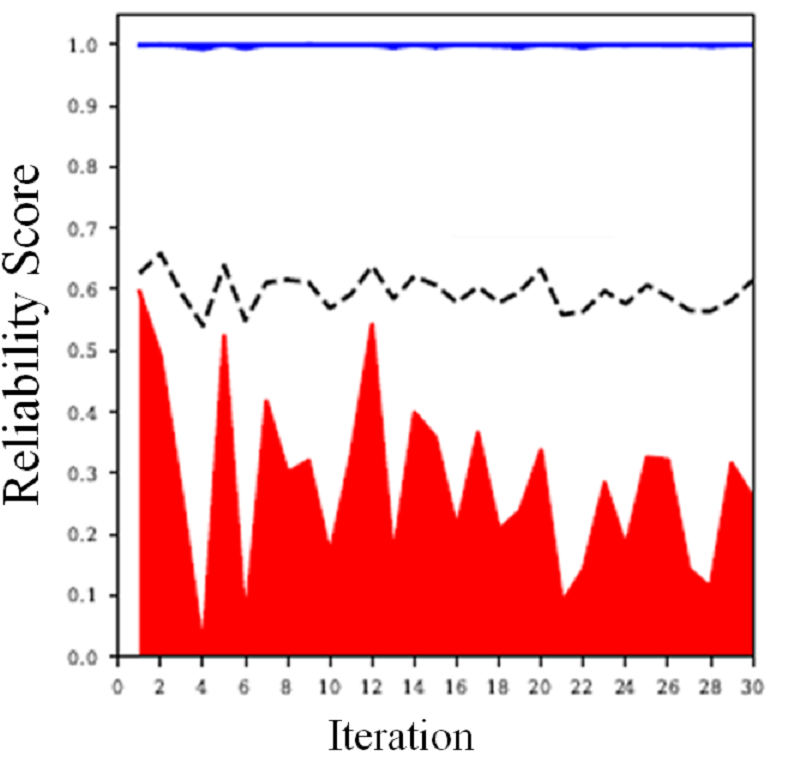}
\caption{Byzantine($RobustFed$)}
\end{subfigure}%
\begin{subfigure}{0.28\textwidth}
\includegraphics[width=\textwidth]{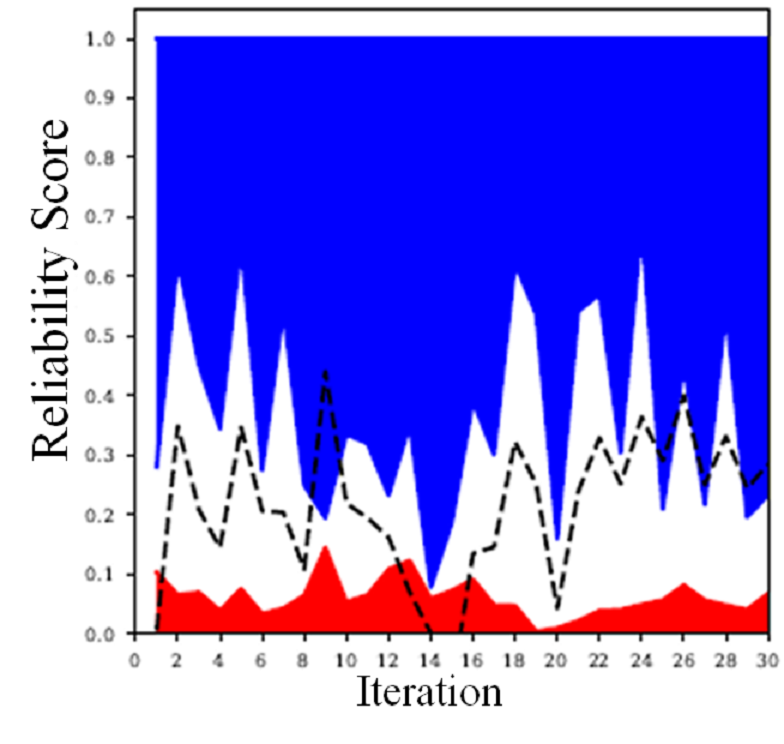}
\caption{Flipping($RobustFed_{+}$)}
\end{subfigure}%
\begin{subfigure}{0.28\textwidth}
\includegraphics[width=\textwidth]{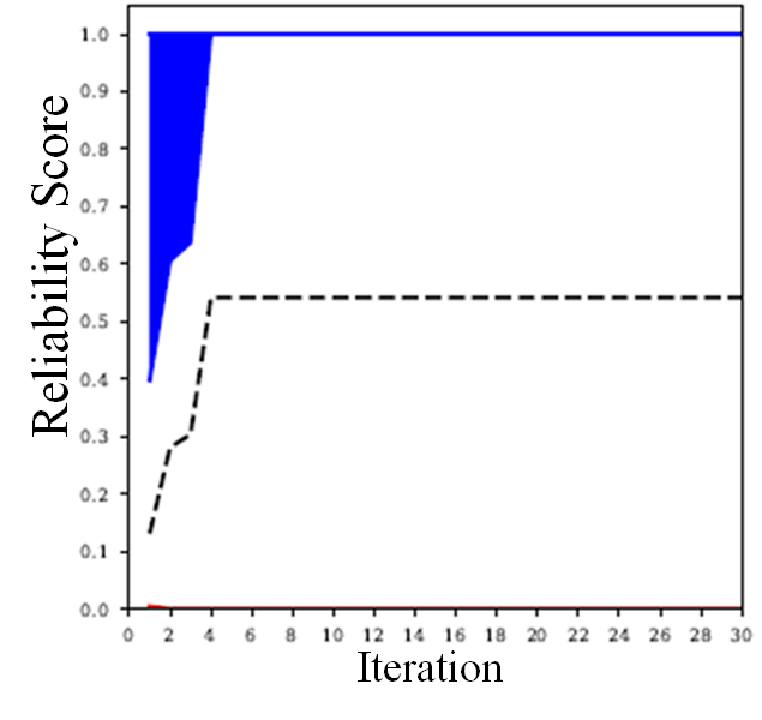}
\caption{Byzantine($RobustFed_{+}$)}
\end{subfigure}%

\caption{Range of Clients' Reliability on FMNIST dataset (10 clients, 30\% malicious clients) }
\label{fig:attack_rel_fmnist}
\end{figure}

\subsection{Experiment Results}
\partitle{Effect of Attacks on Reliability Score of Clients}
Figure~\ref{fig:attack_rel_fmnist} shows the reliability range of malicious and benign clients under label-flipping and Byzantine attacks in static mode learned by RobustFed and RobustFed$_t$, correspondingly. 
We observe that RobustFed assigns higher reliability to benign workers and vice versa under Byzantine attack and noisy data attack as we expected. However, the opposite behavior is observed under flipping attack. As we discussed, this is likely because the gradients of the malicious clients are outliers under such attacks and significantly dominates (biases) the aggregated model parameters, and hence has high reliability due to the Euclidean distance based evaluation. Therefore, in our Robust$_+$ approach, we disregard the clients with both high or low reliabilities, which will help mitigate the impact of the malicious clients.


For Robust$_t$, by incorporating the statistical information of previous rounds, it is able to correctly assign higher reliability to the benign clients (even though with some fluctuations under flipping attacks). It's worth noting that it separates the two types of clients extremely well under Byzantine attack and successfully recognizes malicious clients in all attacks, i.e., assigning close to 0 reliability for them. 
\vspace{-1em}

\begin{figure}[]
\centering
\includegraphics[width=0.5\textwidth]{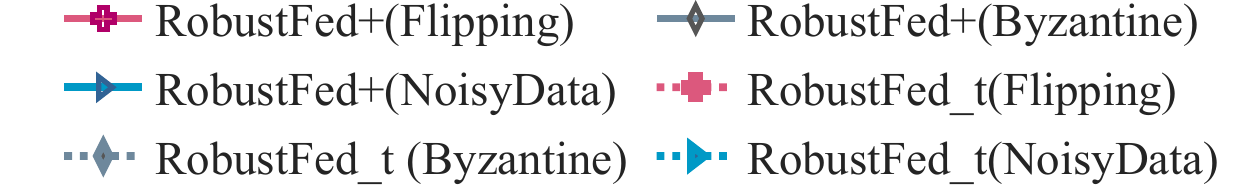}

\includegraphics[width=0.34\textwidth]{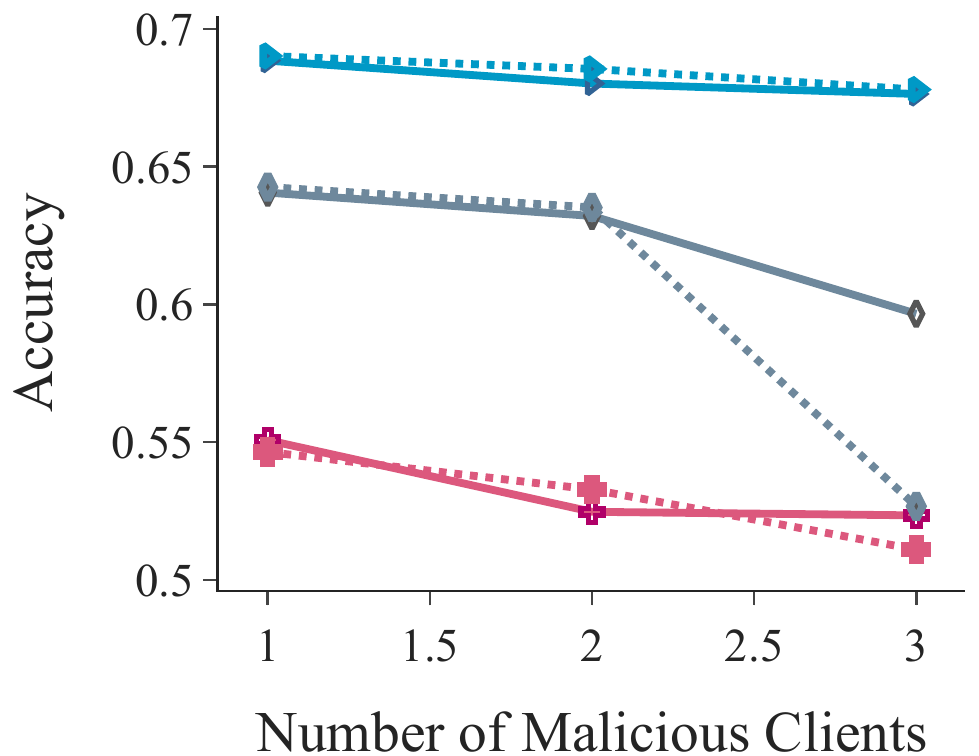}
\includegraphics[width=0.32\textwidth]{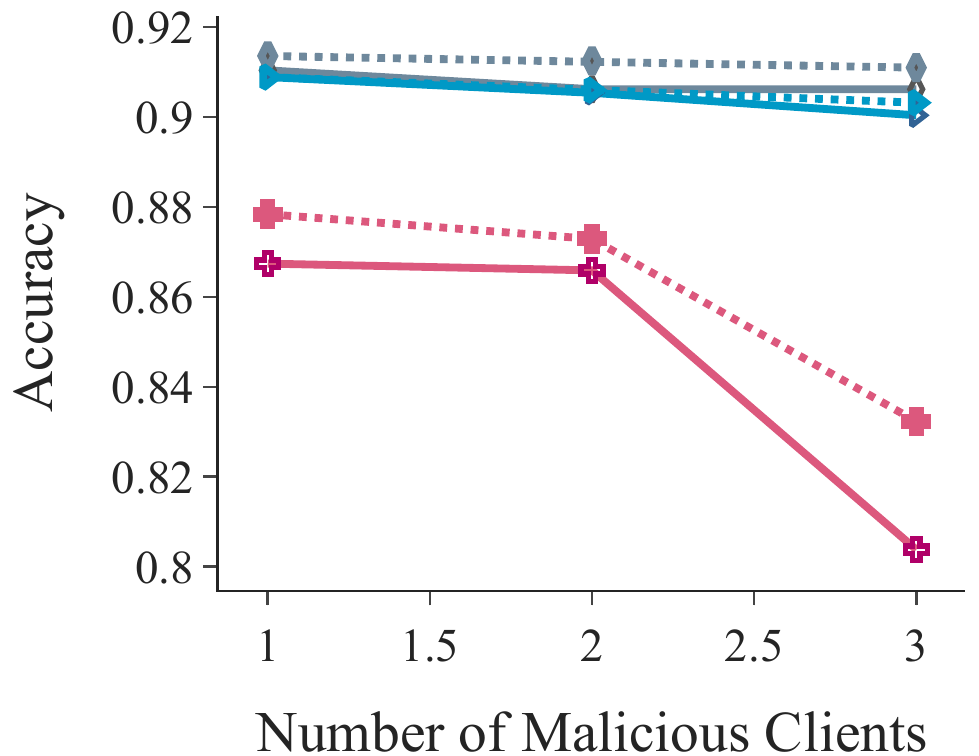}
\includegraphics[width=0.32\textwidth]{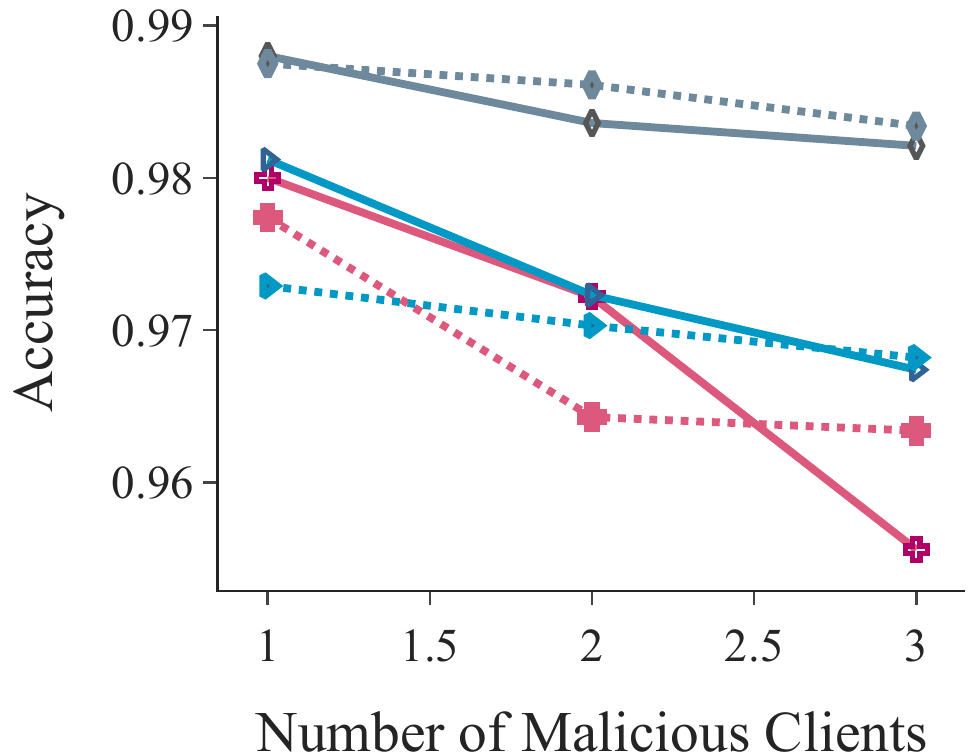}
\caption{Effect of number of Malicious Clients}
\label{fig:attack_rel}
\end{figure}

\partitle{Impact of number of Malicious Clients}
We study the impact of the number of malicious clients on the proposed aggregation method.
As it is shown in Fig.\ref{fig:attack_rel}, 
By increasing the number of malicious clients, the performance of the global model slightly drops. It can be observed that RobustFed$_t$ improves upon RobustFed$_+$ for FMNIST and MNIST datasets that have a higher accuracy on their clean data (i.e., no attack). However, in the CIFAR\_10 dataset that has a poor performance on clean data, RobustFed$_t$ could not improve the performance.

\partitle{Robustness}
In this experiment we compare our robust aggregation methods (RobustFed, RobustFed$_+$, RobustFed$_t$) with the state-of-the-art baselines. The results of these methods along with average performance are shown in Table~\ref{tbl:static_results}.  

\begin{table}[]
\caption{Aggregation Method Comparison in Static \& Dynamic Mode \\ (30\% malicious clients)}
\resizebox{0.9\columnwidth}{!}{%
\begin{tabular}{ |c|c|c|c|c|c|c|c|c|c| } 
\hline
\multicolumn{9}{c}{Static Mode} \\
\hline
Dataset&Attack & FedAvg & Median & Trim\_mean & Krum & RobustFed & RobustFed$_+$ & RobustFed$_t$ \\
\hline \hline
\multirow{1}{*}{CIFAR\_10}& Clean & 70.25 & \bf{70.75} & \bf{70.78}&57.75 & 68.05& 69.74&69.75 \\ 
&Byzantine & 10.0 & 55.01 & 10.29&57.24 & 44.64& \bf{59.66}& 54.67\\
&Flip Label& 51.37 & 41.34 & 46.74 & 10.0 & 10.0 & \bf{52.34}& 51.10\\ 
&Noisy& 67.51 & \bf{68.31} & 68.22 & 57.67 & 67.22 & 67.64& 67.80 \\ \hline 
&Average Performance& 42.96 & 54.88 &41.75& 41.63&40.62 & \bf{59.88}& \bf{58.19}\\ \hline \hline
\multirow{1}{*}{FMNIST}& Clean & \bf{91.15} & 90.95 & 91.05 & 87.79 & 91.05 & 91.05 & 91.07\\ 
&Byzantine & 10.0 & 89.20 & 10.0& 87.66 & 81.25& \bf{90.62}& 84.59\\
&Flip Label& 79.05 & 77.58 & 73.23 & 10.0 & 14.55 & 80.38 & \bf{83.52}\\ 
&Noisy& 89.25 & 89.20 & 89.32 & 84.78 & 84.09 & 87.74& \bf{89.0} \\ \hline
&Average Performance& 59.433 & 85.32 &57.51& 60.81&59.96 & \bf{85.9}& \bf{85.7}\\ \hline \hline
\multirow{1}{*}{MNIST}& Clean & 99.29 & \bf{99.31} & \bf{99.34} & 98.51 & 99.01 & \bf{99.3} &\bf{99.32}\\ 
&Byzantine & 11.35 & 98.18 & 11.35 & 97.43 & 91.35 & 98.21 & \bf{98.34}\\
&Flip Label& 94.58 & \bf{97.80} & 94.47 & 11.35 & 11.40 & 95.56 & 96.34\\ 
&Noisy& 92.08 & 93.01 & 88.26 & 83.16 & 80.04 & 96.74 & \bf{96.82} \\ \hline
&Average Performance& 66 & 96.33 &64.69& 63.98 & 60.93 & 96.8& \bf{97.2}\\ \hline

\hline
\multicolumn{9}{c}{Dynamic Mode} \\
\hline
Dataset&Attack & FedAvg & Median & Trim\_mean & Krum & RobustFed & RobustFed$_+$ & RobustFed$_t$ \\
\hline \hline
\multirow{1}{*}{CIFAR\_10}& Clean & 69.22 & \bf{69.58} & 68.22&56.69 & 67.87& \bf{69.22}&67.25 \\ 
&Byzantine & 12.53 & 44.93 & 10.00 &\bf{61.49} & 55.0& 58.78& \bf{60.56}\\
&Flip Label& 10.0 & 35.00 & 10.07 & 10.32 & 11.56 & \bf{57.73}& 55.53\\ 
&Noisy& 63.27 & 63.35 & 61.18 & 61.36 & 61.67 & 63.43& \bf{63.78} \\ \hline
&Average Performance& 28.6 & 47.76 &27.08& 44.39& 42.74 & \bf{59.98}& \bf{56}\\ \hline \hline
\multirow{1}{*}{FMNIST}& Clean & 91.68 & 92.00 & 88.26 & 89.79 & 91.79 & 91.98 & 91.87\\ 
&Byzantine & 10.0 & 88.90 & 25.0& \bf{90.36} & 81.35& \bf{89.85}& 83.00\\
&Flip Label& 10.0 & 68.23 & 10.25 & 11.04 & 11.35 & 70.93 & \bf{78.24}\\ 
&Noisy& 89.08 & 88.12 & 86.13 & 81.12 & 89.24 & 90.01 & \bf{90.24} \\ \hline
&Average Performance& 36.36 & 81.75 & 40.46 & 60.84 & 60.64 & 83.49 & \bf{83.82}
\\ \hline \hline 
\multirow{1}{*}{MNIST}& Clean & 99.32 & 99.35 & 99.28 & 99.01 & 99.32 & 99.34 & 99.33\\ 
&Byzantine & 11.35 & 97.05 & 10.01& 96.37 & 96.27 & 97.07 & 94.38\\ 
&Flip Label&10.28 & 94.63 & 10.54 & 11.35 & 12.16 & 94.99 & 95.23\\ 
&Noisy& 80.12 & 96.67 & 95.34 & 94.23 & 87.37 &96.10 & 96.07 \\ \hline
&Average Performance&33.91&95.95&38.63&67.31&65.26&\bf{96.05}&95.22

\\ \hline 

\end{tabular}
}
\label{tbl:static_results}
\end{table}

\begin{itemize}
\item{\bf{Static Mode.}}

In this experiment, clients that participate in each round are fixed. The total number of clients are considered to be 10, in which 30\% of them (i.e., 3 clients) are malicious ones. 
As shown in Table~\ref{tbl:static_results}, RobustFed$_+$ and RobustFed$_t$ provide more consistent and better robustness against all three types of attacks while having comparable accuracy on clean data compared with all state-of-the-art methods. As expected, FedAvg’s performance is significantly affected under the presence of malicious clients, especially in Byzantine and flipping attacks. It is also interesting to observe that both Krum and Median are very sensitive to label flipping attacks. 
\item{\bf{Dynamic Mode.}} In this experiment, at each round, 10 clients are randomly selected from a pool of 100 clients consists of 30 malicious clients and 70 normal clients. 
We observe that RobustFed$_+$ performs stronger robustness by incorporating historical information. 
\end{itemize}

\section{Conclusions \& Future Works}
In this paper, we have studied the vulnerability of the conventional aggregation methods in FL. We proposed a truth inference approach to estimate and incorporate the reliability of each client in the aggregation, which provides a more robust estimate of the global model. In addition, the enhanced approach with historical statistics further improves the robustness. Our experiments on three real-world datasets show that RobustFed$_+$ and RobustFed$_t$ are robust to malicious clients with label flipping, noisy data, and Byzantine attacks compared to the conventional and state-of-the-art aggregation methods. This study focuses on data with IID distribution among clients; future research could consider non-IID distribution. 


\bibliographystyle{acm}
\bibliography{paper}

\end{document}